\documentclass[runningheads]{llncs}

\usepackage{eccv}

\usepackage{eccvabbrv}

\usepackage{graphicx}
\usepackage{booktabs}

\usepackage[accsupp]{axessibility}  

\usepackage{hyperref}

\usepackage{orcidlink}

\usepackage{times}
\usepackage{soul}
\usepackage{url}
\usepackage{graphicx}
\usepackage{mathtools}
\usepackage{amsmath}
\usepackage{amssymb}
\usepackage{bm}
\usepackage{nicematrix}
\usepackage{booktabs}
\usepackage{algorithm}
\usepackage{algorithmic}
\usepackage{colortbl}
\usepackage{multirow}
\usepackage{wrapfig}

\begin{document}

\title{Geometrically Constrained and Token-Based \\Probabilistic Spatial Transformers} 

\titlerunning{TokenSTN}

\author{Johann Schmidt\inst{1}\orcidlink{0000-0002-9700-3069} \and
Tom Siegl\inst{2}\orcidlink{0000-0003-2292-1188} \and
Martin Becker\inst{3}\orcidlink{0000-0003-4296-3481} \and
Sebastian Stober\inst{1}\orcidlink{0000-0002-1717-4133}}

\authorrunning{Schmidt et al.}

\institute{Artificial Intelligence Lab, Otto-von-Guericke University Magdeburg, Germany
\email{\{johann.schmidt, stober\}@ovgu.de} \and
University of Rostock, Germany \email{tom.siegl@uni-rostock.de} \and
Becker Lab, University of Marburg, Germany \email{martin.becker@uni-marburg.de}}

\maketitle

\begin{abstract} 
Spatial transformations such as rotation and scale obscure the morphological cues needed for accurate image classification. 
Careful consideration is required for reliable use in high-stakes settings. 
A model should stay robust under such transformations, expose why a correction was applied, and signal when its input is ambiguous. 
While geometrically equivariant architectures provide a mathematically grounded solution, they often limit model flexibility through strict symmetry constraints and incur significant computational overhead. 
Spatial Transformer Networks (STNs) offer a data-driven, flexible alternative for learning pseudo-equivariances to affine transformations. 
However, STNs have historically been restricted to convolutional architectures and suffer from training instability.
To address this, we introduce a novel STN framework.  
It leverages the global modeling capabilities of transformers to regress the affine transformation acting on the input. 
For this, we decompose affine transformations into interpretable primitives, regressed under adaptable geometric constraints, thereby preventing the training instability typically caused by degenerate transformations. 
By sharing weights between the localization network and the classification backbone, the framework requires minimal computational overhead.
Extensive experiments on challenging insect biodiversity and medical imaging benchmarks demonstrate that our approach achieves superior predictive performance under diverse spatial transformations while maintaining high efficiency.
Code is available at \url{https://github.com/johSchm/TokenSTN}.
\end{abstract}

\begin{figure}[h!]
    \centering
    \includegraphics[width=0.7\linewidth]{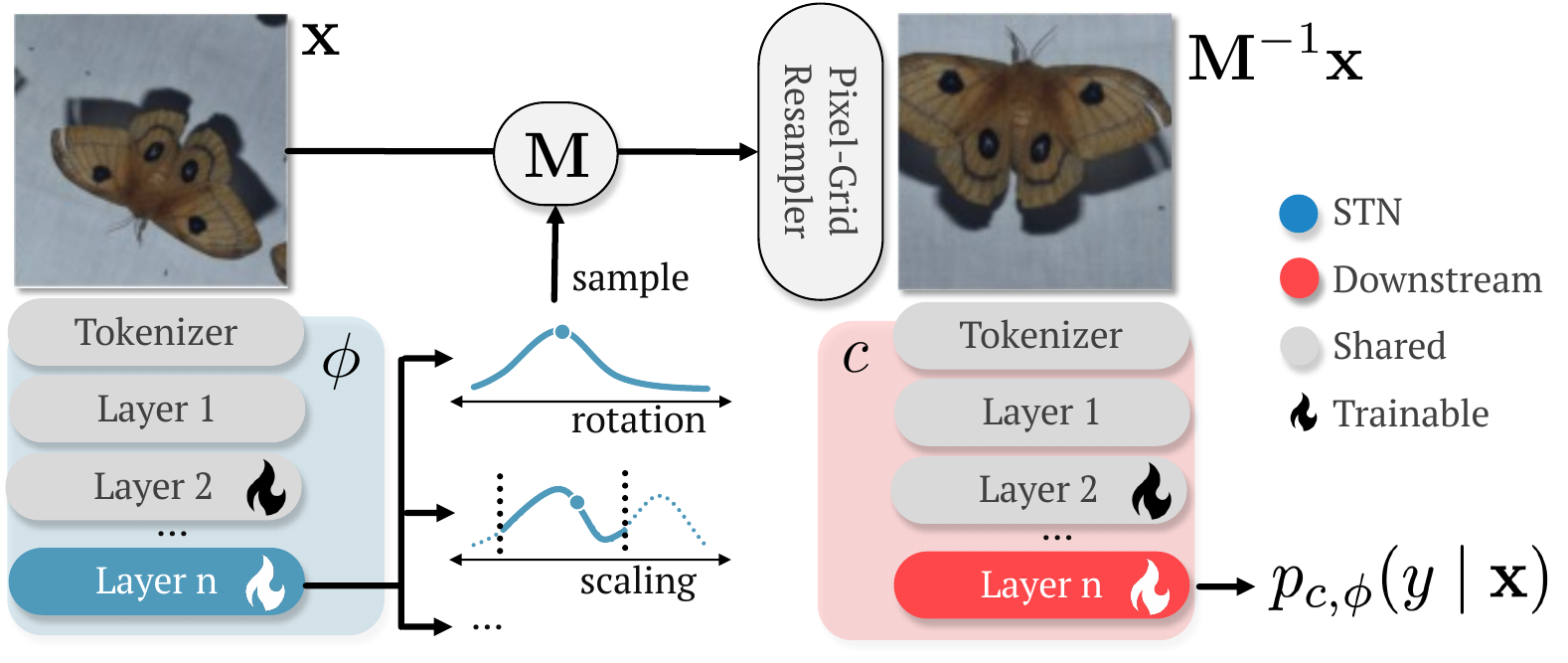}
    \caption{
        Our method integrates a Spatial Transformer module directly into the Vision Transformer pipeline. Unlike traditional STNs, we regress geometric primitives (e.g., rotation, scale) individually under adaptable constraints to ensure training stability and prevent degenerate transformations. By sharing weights between the localization network and the classification backbone, the framework eliminates feature redundancy and maintains high computational efficiency.
    }
    \label{fig:overview}
\end{figure}

\section{Introduction} \label{sec:introduction}

\paragraph{Vision Models are Brittle}
Deep vision models exhibit notorious brittleness when facing spatial variations \cite{Engstrom2019}. 
Because distinct spatial transformations of a single object project onto vastly different points in the pixel manifold, the resulting signals can disrupt prediction consistency. 
As illustrated in \cref{fig:overview}, while both images depict the same moth, the right, rectified version presents a canonical pose that is intuitively easier to label. 
Unlike humans, who naturally prefer such rectified views, neural networks must infer this preference entirely from their training distribution. 
Consequently, classifiers that view these transformed versions as distinct inputs often fail to learn consistent label functions, necessitating explicit mechanisms to ensure robustness against spatial deformation \cite{Engstrom2019}.

\paragraph{Limitations of Data Augmentation and Invariant Models}
A theoretically grounded solution is the use of inductive biases to render the classifier inherently invariant—ensuring the model predicts the same class regardless of the input's transformation. 
Convolutional Neural Networks gain translation invariance through spatial pooling \cite{BlurPool}, but remain highly sensitive to rotation and scale. 
To achieve broader invariance, filters can be lifted to a filter bank over which pooling is applied \cite{GCNN}, or the predictions of any model over all (or many) transformed inputs can be pooled \cite{Puny2022}. 
However, strictly invariant methods often trade flexibility for efficiency. 
They face significant limitations regarding computational cost, increased memory footprints, and the difficulty of engineering constraints for complex transformation groups \cite{Steerable}.
Distinct from strict invariance, pseudo-invariant methods achieve robustness in a data-driven fashion, a paradigm that has driven recent breakthroughs in computer vision \cite{Tu2024}.
The simplest solution is to augment the classifier's input with transformed input samples.
However, this forces the classifier to either learn replicated features for every transformation \cite{olah2020} or learn ``fuzzy'' averages that obscure detail \cite{Chen2019InvarianceRV, kernelDA2019}.

\paragraph{Spatial Transformer Networks}
A more efficient data-driven alternative is the Spatial Transformer Network (STN) \cite{STN}, which decomposes geometric variance handling (localization network) from the classification task (downstream classifier). 
Acting as differentiable rectifiers, they employ a localization network to regress affine transformation parameters to minimize the downstream classification objective.
This provides a mechanism for learning input-dependent rectifications without introducing the architectural rigidity of invariant models, allowing the downstream classifier to focus purely on the rectified features.
While there has been a long line of further developments \cite{Lin2017, Shu2018, Schwobel2022}, some open problems still prevail.
STNs are notoriously difficult to train due to training instability and computational redundancy. 
The standard approach involves regressing an entire affine matrix \cite{STN}. For training STNs without explicit constraints, the optimization landscape is riddled with degenerate solutions, such as collapsing the image into a line.
As a result, training becomes unstable because degenerate solutions eliminate semantic information, leaving the classifier to operate on effectively meaningless inputs.
One solution is to strictly enforce invertibility \cite{Detlefsen2018}, which increases computational cost and complicates training due to additional integration steps and regularization terms.
Another is to limit the regression to specific subgroups like rotation \cite{Tuggener2023, Schmidt2023}, which are naturally invertible.
This, however, sacrifices the geometric flexibility required for different applications.
Regressing affine transformations depends on the classifier's inferred preference from the training distribution \cite{Boostrapping} and the object class itself.
This can lead to computational redundancy, as effective rectification requires semantic understanding, forcing the localization network to duplicate the feature extraction effort of the downstream classifier.
As a solution, \cite{Finnveden2020} proposed different weight-sharing schemes between networks.
These approaches are limited to Convolutional Neural Networks, while transformers (dominating computer vision) have not been used to the best of our knowledge. 
A final issue is the regression of point estimates for transformation parameters, which causes brittle predictions in noisy scenes (common in natural images).
To address this, \cite{Schwobel2022} treats transformation parameters as latent variables with distributions rather than point estimates.
This allows estimating the confidence of a prediction and thus identifying uncertainty.
However, the solution requires a hierarchical prior, which induces a significant engineering effort.

\paragraph{Contributions}
In summary, due to their flexibility STNs have the potential to vastly outperform more restricted specializations.
However, current solutions to make this feasible strongly restrict this flexibility, impact trainability, are computationally expensive, or require substantial engineering effort.
To unlock the full potential of data-driven invariance, we propose a modernized STN framework that reconciles geometric flexibility with training stability\footnote{The stability provided by our bounded regression should be understood with respect to the transformed and resampled signal itself, rather than the gradient directly. Specifically, our bounding strategy prevents degenerate transformation extremes (such as excessive zoom-out that renders the target object invisible) while avoiding the substantial computational overhead associated with diffeomorphic integration \cite{Detlefsen2018}.} and computational efficiency.
Our contributions are:

\begin{itemize}
    \item To prevent the degenerate solutions common in unconstrained STNs without the computational cost of diffeomorphic integration, we introduce a bounded geometric decomposition. By regressing individual primitives (e.g., rotation, scale) under lightweight constraints, we ensure invertible and interpretable transformations.
    \item We extend the STN paradigm to transformers through a weight-sharing scheme between the localization network and the classifier, eliminating architectural redundancy and lifting spatial rectification to state-of-the-art vision architectures.
    \item To improve robustness for noisy scenes, we simplify the variational objective of \cite{Schwobel2022} by employing independent Gaussian posteriors for each transformation primitive, achieving competitive results.

\end{itemize}
\noindent
Together, robustness and geometric interpretability, achieved at minimal cost, target central requirements of trustworthy vision models in high-stakes domains.
We demonstrate improved robustness against affine transformed inputs for biodiversity monitoring \cite{EU} and medical diagnostics \cite{medMNIST}.

\section{Background} \label{sec:affine}
Classification on a dataset $\mathcal{X}$ often involves transformations that preserve class identity. 
For instance, an insect rotated by $90^\circ$ is still the same insect. 
As these images have the same label, they belong to the same \emph{equivalence class}.
A robust classifier provides one label per equivalence class.
To study prediction-robustness against such transformations formally, group theory is often used \cite{ITS}. 
We introduce only the necessary vocabulary used in this paper and refer for more details to \cite{bronstein2021geometric, hall2015lie}.
Let $G$ be a group, like the group of affine transformations $\mathrm{Aff}_2(\mathbb{R})$ of planar transformations in 2D.
A \emph{group} $G$ is a set equipped with an associative binary operation, an identity element, and inverses.
A group element $g \in G$ can act on an image $\mathbf{x}$ by transforming it.
Group elements are abstract objects, so a representation is needed to perform such action.
Affine transformations are represented by some invertible matrix $\mathbf{M} \in \mathbb{R}^{3 \times 3}$ (using homogeneous coordinates).
Generally, we only use matrix Lie groups \cite{hall2015lie} in this paper, hence all representations are matrices.
A subgroup of a group $G$ is a subset $G^\prime \subseteq G$ that is itself a group under the operation inherited from $G$.
The \emph{orbit} of $\mathbf{x}\in\mathcal{X}$ is $G\mathbf{x} = \{ g\mathbf{x} : g\in G\}$, the set of all symmetry-related versions of $\mathbf{x}$. 

In this work, we are interested in the process of \emph{canonicalization}, which selects a concrete representative from each orbit.
Canonicalizers are modules, which can be plugged in front of any downstream model, hence they are agnostic to it. 
This allows us to train a canonicalizer and apply it to any pre-trained downstream model, like a large vision classifier, without the cost of re-training.
Formally, a \emph{canonicalizer} $\phi$ is a map $\phi:\mathcal{X}\to G$ such that $\phi(g\mathbf{x})^{-1} g \mathbf{x} = \mathbf{x}$ for all $g\in G$, and $\phi(\mathbf{x})\mathbf{x} \in G\mathbf{x}$ lies in the same orbit as $\mathbf{x}$. 
That is given a transformed image $\mathbf{M} \mathbf{x}$, we undo the transformation by $\phi(\mathbf{M} \mathbf{x})\mathbf{M} \mathbf{x} = \mathbf{M}^{-1} \mathbf{M} \mathbf{x} = \mathbf{x}$.
Existing approaches to obtain such functions are limited to equivariant backbones \cite{Kaba2023, Mondal2023}.
We revisit STNs as pseudo-canonicalizers (we say rectifiers instead), which come without such hard and costly constraints but also do not mathematically guarantee equivariance.
STNs use a (unconstrained) localization network $\phi:\mathcal{X}\to \mathbb{R}^6$ to learn a data-point dependent parameterization of $\mathbf{M}$.
We define $\mathcal{T} \in \mathbb{R}^6$ as the set of parameters to define $\mathbf{M}$.
A \emph{grid generator} transforms the regular pixel grid by $\mathbf{M}$ and a \emph{sampler} resamples $\mathbf{x}$ onto this grid via interpolation.
This mechanism is fully differentiable and can in principle be attached to any backbone architecture \cite{STN}. 

\section{Decomposition of Affine Transformations} \label{sec:decomposition}
Vanilla STNs regress the entries of $\mathbf{M}$ directly \cite{STN}. 
However, the raw matrix entries are entangled: a change in one entry can simultaneously affect rotation, shear, and scaling, complicating optimization and interpretability. 
To disentangle these factors, we can utilize the fact that the affine group admits \emph{canonical coordinate} decompositions. 
Canonical coordinates parameterize group elements via simpler subgroups, effectively mapping complex group actions to disentangled operations \cite{PolarTransformer, Tai2019}.
To improve the stability and interpretability of STNs, we regress these canonical coordinates (the parameters of the subgroups of $\mathrm{Aff}_2(\mathbb{R})$) rather than the raw matrix entries. 
This allows individual geometric primitives to be explicitly disabled or bounded. 
While this is often achieved by altering the input domain (e.g., Log-Polar coordinates converting rotation to translation \cite{PolarTransformer}), we enforce this structure directly on the transformation group.
We provide an extension to the Iwasawa decomposition \cite{Iwasawa1949}, \cite[Chapter 6]{Knapp1996} for $\mathrm{Aff}_2(\mathbb{R})$ with explicit reflection and translation primitives.

\begin{lemma}\label{theorem:decomposition}
    Let $\mathbf{M} \in \mathrm{Aff}_2(\mathbb{R})$, which is applied to some vector $\mathbf{x} \in \mathbb{R}^3$ with homogeneous coordinates via left group action such that
    \begin{equation}
        \mathbf{M} \mathbf{x}
        = \begin{bNiceMatrix}[margin]
        \Block{2-2}{\mathbf{A}} &  & \Block{2-1}{\mathbf{t}} \\
        & & \\
        0 & 0 & 1
    \end{bNiceMatrix} \mathbf{x},
    \end{equation}
    with $\mathbf{A} \in \mathrm{GL}_2(\mathbb{R})$ and $\mathbf{t} \in \mathbb{R}^2$.
    $\mathbf{M}$ decomposes \emph{uniquely} into primitive geometric transformations:
    \begin{align}
    \small
    \mathbf{M} =
    \underbrace{\begin{bNiceMatrix}[margin]
        1 & 0 & t_x \\
        0 & 1 & t_y \\
        0 & 0 & 1
    \end{bNiceMatrix}}_{\text{Translation}\; \mathbf{T}}
    \cdot
    \underbrace{\begin{bNiceMatrix}[margin]
        1 & 0 & 0 \\
        0 & r & 0 \\
        0 & 0 & 1
    \end{bNiceMatrix}}_{\text{Reflection}\; \mathbf{F}}
    \cdot
    \underbrace{\begin{bNiceMatrix}[margin]
        \cos\theta & -\sin\theta & 0 \\
        \sin\theta & \cos\theta & 0 \\
        0 & 0 & 1
    \end{bNiceMatrix}}_{\text{Rotation}\; \mathbf{R}}
    \cdot
    \underbrace{\begin{bNiceMatrix}[margin]
        1 & h & 0 \\
        0 & 1 & 0 \\
        0 & 0 & 1
    \end{bNiceMatrix}}_{\text{Shear}\; \mathbf{H}}
    \cdot
    \underbrace{\begin{bNiceMatrix}[margin]
        s_x & 0 & 0 \\
        0 & s_y & 0 \\
        0 & 0 & 1
    \end{bNiceMatrix}}_{\text{Scaling}\; \mathbf{S}}.
    \end{align}
    Here, $\mathbf{t} = (t_x, t_y)^\top$ is the translation vector, $r \in \{1, -1\}$ is the canonical reflection parameter, $\theta \in [0, 2\pi)$ is the rotation angle, $h \in \mathbb{R}$ is the canonical shear parameter, and $s_x, s_y > 0$ are the scaling factors.
\end{lemma}

The proof follows in the Supplementary Material.
This decomposition is particularly advantageous in neural networks as it enforces geometric constraints naturally: rotation preserves orthogonality and area, scaling maintains axis alignment while changing area proportionally, shearing preserves area but distorts angles, and each parameter has a clear geometric interpretation, facilitating both optimization and interpretability in spatial transformer networks.
This decomposition preserves the full degrees of freedom of $\mathrm{Aff}_2(\mathbb{R})$, which renders solutions unique.
Uniqueness in a decomposition means that there is exactly one way (up to any explicitly allowed equivalences, such as ordering or sign) to write an object as a combination of specified components.
As this decomposition is unique (see the Supplementary Material), there is exactly one solution (up to any explicitly allowed equivalences, such as ordering or sign) to express $\mathbf{M}$ by its factors $\mathbf{T} \mathbf{F} \mathbf{R} \mathbf{H} \mathbf{S}$.
Our aim is to regress the parameters of each geometric primitive.
To ease the problem further, we leverage a common approach in computer vision \cite{Lindeberg1998}: That is, we enforce \emph{isotropic scaling} ($s_x = s_y = s$) to preserve the aspect ratio of the object.\footnote{
Anisotropic scaling is not exactly representable under this constraint.
$\operatorname{diag}(s_x,s_y)$ with $s_x\neq s_y$ is not a shear composed with isotropic scaling, and combined with a rotation it cannot be captured at all. 
The model can match the anisotropy magnitude through $h$ but not its orientation.
}
Under this constraint the scaling matrix simplifies to $\mathbf{S} = \operatorname{diag}(s,s,1)$.
This reduces the degrees of freedom from six to five, which prevents the exact representation of arbitrary affine maps but it suffices for the transformations of interest, which do not rescale the axes independently.

\section{STN Framework} \label{sec:methodology}
\begin{wrapfigure}[30]{r}{0.5\textwidth}
    \centering
    \vspace{-1.7cm}
    \includegraphics[width=1.0\linewidth]{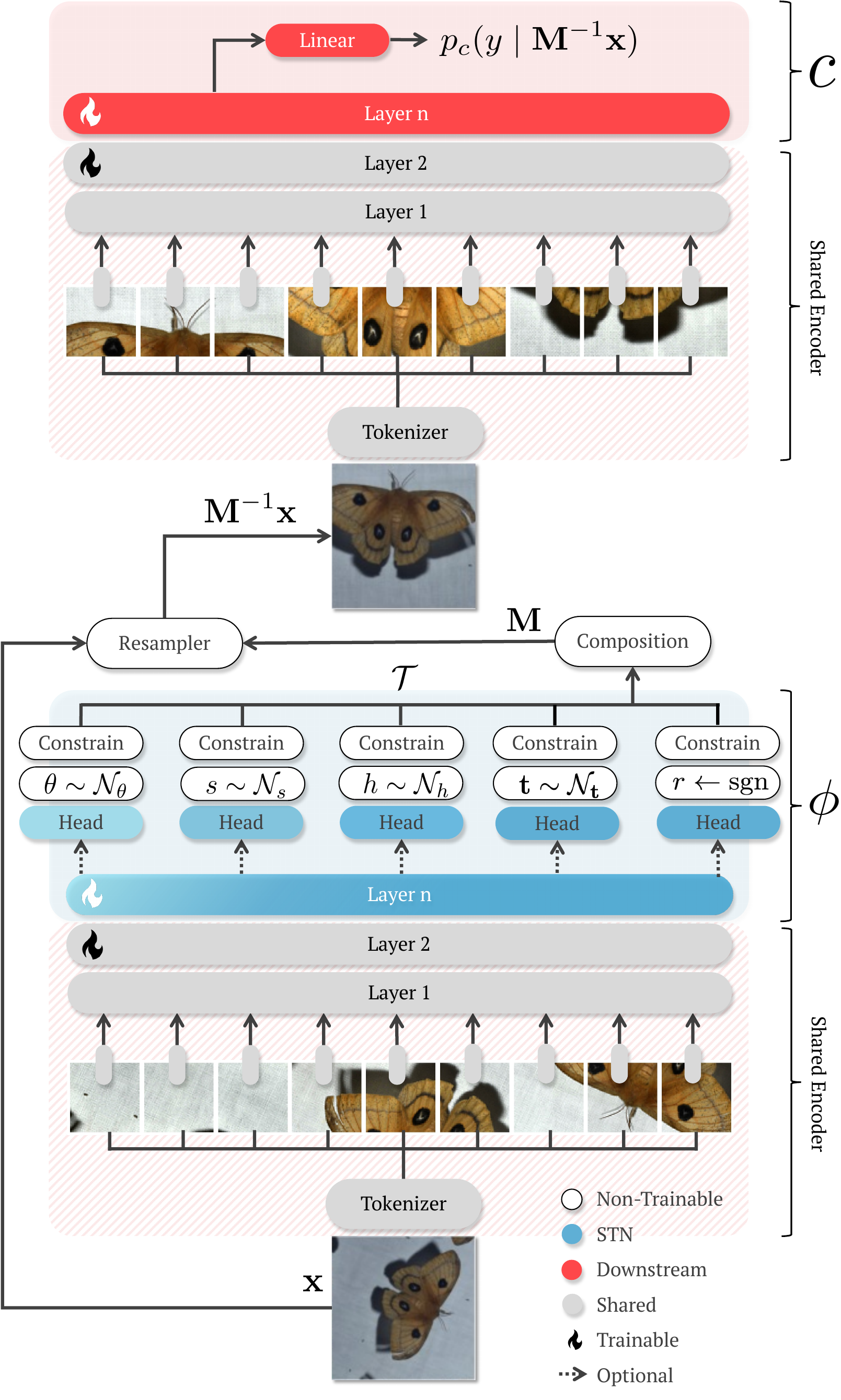}
    \caption{
        Our model first encodes the input image into patch tokens using a frozen backbone tokenizer. 
        A shared localization encoder processes these tokens, and separate regression heads predict affine transformation parameters for rotation, scaling, shearing, translation, and reflection (each is optional).
        The resulting affine matrix rectifies the image, which is then re-tokenized and fed into a downstream classifier.
    }
    \label{fig:arch}
    \vspace{-1.5cm}
\end{wrapfigure}

Our methodology proceeds by jointly encoding the input to extract task-relevant features, regressing a set of constrained geometric primitives to parameterize an affine transformation, and finally resampling the image into a rectified pose for downstream classification.
Let $(\mathbf{x}, y) \sim \mathcal{D}$ represent a sampled tuple from a dataset $\mathcal{D}$, where $\mathbf{x} \in \mathbb{R}^{C \times H \times W}$ is an image and $y \in \mathcal{Y} \subset \mathbb{N}$ is its corresponding class label.
Here, $C \in \mathbb{N} \setminus \{ 0 \}$ is the number of channels, usually $C=3$ for RGB images, and $H,W \in \mathbb{N} \setminus \{ 0 \}$ the spatial dimensions of the image. 
In \cref{fig:arch} we provide an overview of our proposed pipeline.
In the following we introduce and discuss its core building blocks.

\subsection{Encoders}
Let $\phi: \mathcal{X} \to \mathcal{T}$ be the localization network used for the regression of $\mathbf{M}$ and $c: \mathcal{X} \to \mathcal{Y}$ be the downstream classifier.
To prevent the complex and redundant feature extraction from pixel space to transformation space, we build on \cite{Finnveden2020} and leverage a shared encoder for $c$ and $\phi$ (see the Supplementary Material for measurements on computational cost).
We show how this can be extended to token-based architectures. 

\paragraph{Shared Encoder}
We share the (frozen) tokenizer and initial layers of $c$ with $\phi$.
The shared encoder maps images to $K$ token embeddings.
Since the localization network $\phi$ outputs rectified images the same encoder can be used for the downstream classifier $c$ without architectural modifications.
This weight-shared design facilitates a backbone-agnostic interface while enforcing a task-informed functional co-dependency that semantically aligns spatial rectification with the downstream classification objective.
This allows for flexible deployment across diverse vision applications while preserving the benefits of pre-trained representations.

\paragraph{Localization Encoder}
We further encode the $K$ tokens from the shared encoder by a localization encoder $\phi$.
The shared tokens comprise low-level task-specific semantics, from which $\phi$ extracts high-frequency geometric information.
With this, the geometric primitives $\mathcal{T}$ can be estimated.
$\phi$ first employs a shallow (2-layer) vision transformer encoder 
resulting in $C$-dimensional latent embeddings for all tokens $\mathbb{R}^{K \times C}$, which can encode high-frequency features and context (like the shape of an object).
Then, $\phi$ applies a permutation-invariant pooling over $K$ (we pool by averaging) $\mathbb{R}^{K \times C} \to \mathbb{R}^{C}$.
For each geometric primitive in $\mathcal{T}$, this pooled representation is further processed by individual regression heads, i.e., a shallow (2-layer) MLP with non-linearities (we use GeLUs).

\subsection{Regression Heads}
We compose the shared localization encoder with individual regression heads to form the localization network $\phi$.
As shown in \cref{theorem:decomposition}, the $\mathrm{Aff}_2(\mathbb{R})$ decomposes into five geometric primitives.
Together with the isotropic scaling assumption in \cref{sec:decomposition}, this gives two translation parameters $(t_x, t_y)$, one scaling parameter $s$, one reflection parameter $r$, one rotation parameter $\theta$ and one shearing parameter $h$.
This parametric decomposition allows us to regress each transformation component separately and compose $\mathbf{M}$ through matrix multiplication.
We use one head architecture with individual weights for each geometric primitive.
All heads consist of a linear layer extracting features from the shared representation $\mathbf{z} \in \mathbb{R}^{C}$.
We initialize each head with zero weights and a bias close to the identity transform, which provides a stable warm start.
Optionally, a primitive can be disabled by freezing its head at this initialization (\cref{fig:arch}).

\paragraph{Probabilistic Regression}
Instead of regressing each geometric parameter as a point estimate, which ignores prediction uncertainty, each head parameterizes a distribution over an unbounded transformation domain and samples from it.
\cite{Schwobel2022} proposed a hierarchical Bayesian model in which a Gamma hyperprior on the precision of a Gaussian latent induces a Student-t marginal posterior over transformations $\mathcal{T}$.
This formulation allows adaptive variance and theoretical robustness to outliers.
However, computing the Kullback--Leibler (KL) divergence for the latent Gamma posterior adds gradient variance and sampling from a hierarchical distribution increases computational overhead.
Instead we simplify the approach by modeling each parameter with a Gaussian posterior and a Gaussian prior.
We show that a simple Gauss distribution $\mathcal{N}(\mu_{\mathcal{T}}, \Sigma_{\mathcal{T}})$ obtains competitive results in our tests.
Each head outputs an unbounded location $\mu_{\mathcal{T}}$, which serves as the distribution mean, and a log-variance $\log \sigma^2$:
\begin{align}
    \mu_{\mathcal{T}} = \mathbf{W}_{\mathcal{T}} \mathbf{z} + \mathbf{b}_{\mathcal{T}}
    \quad \text{and} \quad
    \sigma_{\mathcal{T}} = \exp \left( \tfrac{1}{2} \alpha_{\mathcal{T}} \operatorname{tanh} \left( \bar{\mathbf{W}}_{\mathcal{T}} \mathbf{z} + \bar{\mathbf{b}}_{\mathcal{T}} \right) - \tfrac{1}{2} \beta_{\mathcal{T}} \right), \label{eq:loc_std_head}
\end{align}
where $\mathbf{W}_{\{s, h, \theta, r\}}, \bar{\mathbf{W}}_{\{s, h, \theta, r\}} \in \mathbb{R}^{C \times 1}$ and $\mathbf{W}_{\mathbf{t}}, \bar{\mathbf{W}}_{\mathbf{t}} \in \mathbb{R}^{C \times 2}$, with bias vectors of matching dimension.
The constants $\alpha_{\mathcal{T}}, \beta_{\mathcal{T}} \in (0, \infty)$ bound the log-variance to a reasonable domain.
We have $\log \sigma^2 \in \left[-\alpha-\beta, \alpha-\beta\right]$ and hence a variance range of $\sigma^2 \in \left[\exp(-\alpha-\beta), \exp(\alpha-\beta)\right]$.
We draw a sample in this unbounded space using the reparameterization trick \cite{VAE}
\begin{equation} \label{eq:sampling}
    \tilde{\mu}_{\mathcal{T}} = \mu_{\mathcal{T}} + \sigma_{\mathcal{T}} \odot \mathbf{\epsilon}_{\mathcal{T}}
    \quad \text{with} \quad
    \mathbf{\epsilon}_{\mathcal{T}} \sim \mathcal{N}(\mathbf{0}, \mathbf{I}),
\end{equation}
where $\odot$ denotes the Hadamard product.
As $\tilde{\mu}_{\mathcal{T}}$ is unbounded, $\mathbf{M}$ could  degenerate.

\paragraph{Bounded Regression}
We therefore map each sampled location into the admissible range of its primitive by a component-specific squashing function, which keeps $\mathbf{M}$ invertible (avoiding singular matrices):
\begin{align} \label{eq:bound_heads}
\theta &= \lambda_\theta \pi \operatorname{tanh}\left( \tilde{\mu}_\theta \right) \in \left[-\lambda_\theta\pi, \lambda_\theta\pi\right],
&
s &= 1-\lambda_s \operatorname{sigmoid}\left( \tilde{\mu}_s \right) \in \left[1-\lambda_s, 1\right], \nonumber \\
h &= \lambda_h \pi \operatorname{tanh}\left( \tilde{\mu}_h \right) \in \left[-\lambda_h\pi, \lambda_h\pi\right],
&
\mathbf{t} &= \lambda_t \operatorname{tanh}\left( \tilde{\mu}_{\mathbf{t}} \right) \in \left[-\lambda_t, \lambda_t\right]^2.
\end{align}
where $\lambda_{\{\theta, h\}} \in (0, 1]$, $\lambda_{s} \in (0, 1)$ (to ensure the strict positiveness of $s$), and $\lambda_t \in (0,1)$ are hyperparameters that bound the co-domains.
We sample before bounding so that the Gaussian noise enters in an unconstrained space, while the squashing guarantees a valid transformation for every draw.
Bounding the scale is key, as $\operatorname{det}(\mathbf{S}) = s_x s_y = s^2$ lets the visual area change with the choice of scale, which causes information loss when resampled to a fixed pixel grid.\footnote{This is why we omitted positive scaling (zooming-out).}
Although $\operatorname{det}(\mathbf{H}) = 1$, $\mathbf{H}$ is not orthogonal; it distorts angles and shapes by skewing the coordinate system, again leading to information loss after resampling.
Translations are bounded to a fixed fraction $\lambda_t$ of the image extent.
Together with the reflection padding used during resampling, this keeps the sampling window close to the image domain and avoids large out-of-domain shifts.
Reflection is discrete, so its squashing function is a sign:
$r = \operatorname{sign}(\tilde{\mu}_r) \in \{-1, +1\}$, applied via the diagonal matrix $\mathbf{F} = \operatorname{diag}(1, r)$, which is involutional ($\mathbf{F}^{-1} = \mathbf{F}$) as a reflection must be. 
The sign is not differentiable, so we use a straight-through estimator~\cite{STE}:
the hard sign is taken in the forward pass, while the backward pass uses a clipped-identity surrogate~\cite{Hubara2016}, with unit gradient for $|\tilde{\mu}_r| \le 1$ and zero gradient otherwise. 
Unlike the continuous primitives, $\tilde{\mu}_r$ needs no bounding, since the sign already maps it onto the admissible set $\{-1, +1\}$. 
This keeps $\phi$ end-to-end differentiable and lets it learn the reflection state jointly with the continuous primitives.
In the deterministic variant we omit sampling and apply the squashing of \cref{eq:bound_heads} directly to the locations $\mu_{\mathcal{T}}$, which then serve as the point estimate.

\subsection{Training and Inference}
\label{sec:training}
The STN $\phi$ predicts geometric parameters $\mathcal{T}$ which are used to parameterize the factors in \cref{theorem:decomposition} to obtain $\mathbf{M}$.
The pixel grid is transformed using $\mathbf{M}^{-1}$ and the image $\mathbf{x}$ is resampled using \cite{STN}.
The transformed image is fed into the classifier $c$ to obtain a class prediction.
During training we minimize the expected loss
\begin{align} \label{eq:loss}
    \mathcal{L} &= \mathbb{E}_{\mathcal{T}\sim p_\phi(\cdot \mid \mathbf{x})}\!\Big[
    \mathcal{L}_{\text{NLL}}\!\big(y, \hat{y}\big)
    + \lambda \mathrm{D}_{\mathrm{KL}} \left[
        p_\phi(\mathcal{T} \mid \mathbf{x}) \,\|\, q(\mathcal{T}) \right]
\Big],
\end{align}
where $\hat{y} = c(\mathbf{M}^{-1}\mathbf{x})$ and $N \in \mathbb{N} \setminus \{0\}$ samples are drawn from the posterior $p_\phi(\mathcal{T} \mid \mathbf{x})$ learned by the probabilistic STN.
As in \cite{Schwobel2022} we minimize the KL divergence between the posterior and the prior.
We adopt a component-wise regularization of the transformation parameters, yielding
\begin{equation} \label{eq:dkl}
\mathrm{D}_{\mathrm{KL}} \left[
p_\phi(\mathcal{T} \mid \mathbf{x}) \,\|\, q(\mathcal{T})
\right]
=
\sum_{t \in \mathcal{T}}
\mathrm{D}_{\mathrm{KL}} \left[
p_\phi(t \mid \mathbf{x}) \,\|\, q(t)
\right].
\end{equation}
Each parameter distribution $p_\phi(\cdot \mid \mathbf{x})$ is regularized toward a unit Gaussian prior, with $q(\theta)=q(\mathbf{t})=q(h)=q(s)=\mathcal{N}(\cdot,1)$.
We detach the predicted means from the gradient computation so that the KL regularization acts solely on the uncertainty estimates.
This prevents the model from being biased toward identity transformations (e.g., zero rotation) and allows it to freely infer data-dependent transformation parameters.
For the rotation parameter $\theta$, we define $\mathrm{D}_{\mathrm{KL}}$ using a wrapped Gaussian to respect the periodic nature of the angular domain.
This ensures the divergence remains continuous at the $2\pi$ boundary and provides a topologically consistent scalar for variance regularization within $\mathrm{SO}(2)$.
During inference, we treat the regression of $\mathcal{T}$ as a latent-variable problem over $\mathrm{Aff}_2(\mathbb{R})$.
After training, we obtain a posterior $p_\phi(\mathcal{T}\mid \mathbf{x})$.
The predictive distribution marginalizes over $\mathcal{T}$:
\begin{equation}
    p_{c,\phi}(y \mid \mathbf{x}) = \int p_c\left(y \mid \rho(\mathcal{T})^{-1}\mathbf{x}\right) p_\phi(\mathcal{T}\mid \mathbf{x}) \, \mathrm{d}\mathcal{T}.
\end{equation}
As in \cite{Schwobel2022} we approximate this expectation with $N$ Monte Carlo samples,
\begin{equation}
    p_\phi(y \mid \mathbf{x}) \approx \frac{1}{N}\sum_{n=1}^N p\left(y \mid \mathbf{M}_n^{-1}\mathbf{x}\right)
    \quad \text{with} \quad
    \mathbf{M}_n \sim p_\phi(\mathcal{T}\mid \mathbf{x}).
\end{equation}
This enforces approximate affine invariance by marginalizing transformation uncertainty during both training and inference.

\section{Experiments} 
\label{sec:experiments}

\begin{figure*}[t]
    \centering
    \includegraphics[width=1.\linewidth]{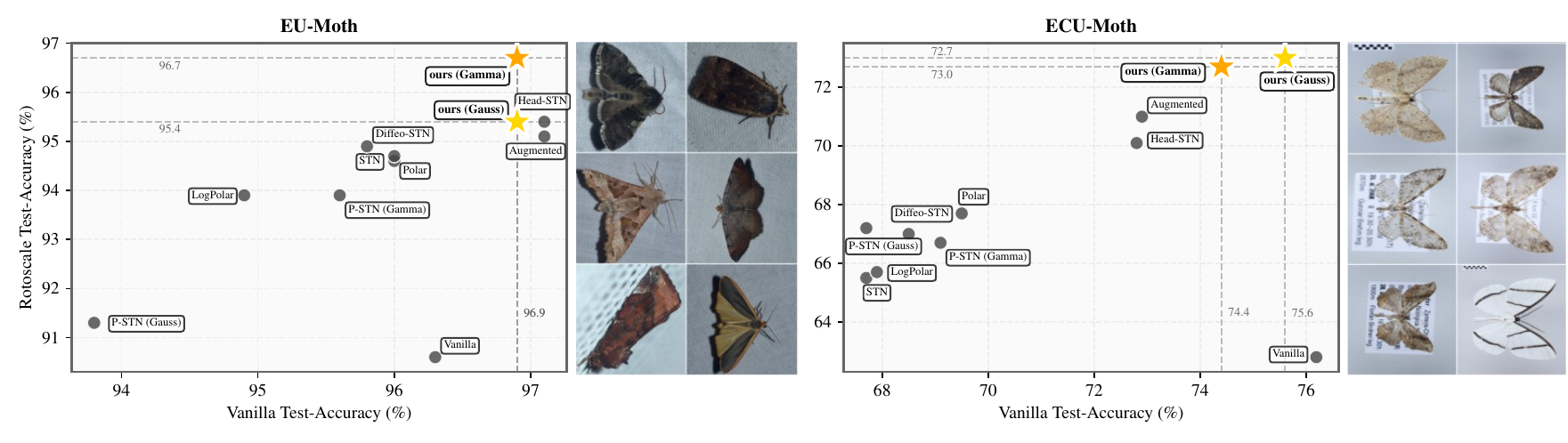}
    \caption{
    Average top-1 accuracy on both moth benchmark test sets versus the rotoscaled test sets. Besides the vanilla SwinB classifier, an augmented version was trained and evaluated and a set of different STNs (dots) including ours (stars).
    }
    \label{fig:moth_performance}
\end{figure*}

\paragraph{Implementation Details}
The experiments demonstrate both performance improvements and increased robustness of our STN compared to the vanilla classifier and other baselines.
For the localization network design we compare our token-based 2-layer transformer encoder with a convolutional network (with batch normalization \cite{BatchNorm}, and adaptive pooling).
For baselines which regress $\mathbf{M}$ without decomposition, we initialize it by identity \cite{STN}, while $\mathcal{T}$ is zero-initialized with identity composition. 
In this paper we focus on the application of biodiversity monitoring of moths \cite{EU,Ecuador} and medical imaging \cite{medMNIST}.
Further details are in our publicly available source code and in the Supplementary Material.

\paragraph{Robustness Evaluation}
We use a principled approach to measure robustness of our model with respect to geometric transformations.
This requires a definition from group theory:
The set of all orbits forms the \emph{orbit space} $\mathcal{X}/G$, which partitions $\mathcal{X}$ into equivalence classes. 
To evaluate the robustness against affine transformed images, we augment the test set $\mathcal{D}_{\text{test}} \subset \mathcal{D}$.
Let $G = \mathbb{R}_{>0} \rtimes \mathrm{SO}(2)$ be the joint group of roto-scaling under matrix multiplication, where $\mathbb{R}_{>0}$ is the space of positive isotropic-scaling.
We construct the orbit space $\mathcal{D}_{\text{test}}/G$.
This holds all roto-scaled variants of the original images in $\mathcal{D}_{\text{test}}$.
In our experiments, we use a finite subgroup $G$ of both $\mathrm{SO(2)}$ and $\mathbb{R}_{>0}$ of order $n=16$.
As we augment images of moths by $G$, we expect only trivial stabilizers, hence low “symmetry richness” in general.
This results in orbit sizes $|G\mathbf{x}| \approx |G|,\;\forall \mathbf{x} \in \mathcal{D}_{\text{test}}$ according to the Orbit-stabilizer theorem \cite{Dummit2004}.
Hence, we expect a large test-time performance drop of models unable to cope with these transformations.

\subsection{Biodiversity Monitoring}
\label{sec:exp:moths}

\begin{wraptable}{r}{0.45\textwidth}
\vspace{-1.2cm}
\centering
\caption{Average top-1 test accuracies on Ecuador-Moth and EU-Moth with shearing augmentation. Relative gains compared to vanilla baseline in parentheses. All methods use pre-trained Swin-Base.}
\label{tab:moths_shearing}
\small
\begin{tabular}{@{}lcc@{}}
\toprule
\textbf{Method} & \textbf{EU} & \textbf{ECU} \\ 
\midrule
Vanilla & 90.2 & 61.7 \\
Augmented & 95.1 {\scriptsize\textcolor{green!70!black}{(+4.9)}} & 70.7 {\scriptsize\textcolor{green!70!black}{(+9.0)}} \\
STN \cite{STN} & 94.6 {\scriptsize\textcolor{green!70!black}{(+4.4)}} & 65.4 {\scriptsize\textcolor{green!70!black}{(+3.7)}} \\
Head-STN \cite{Chen2021} & 95.3 {\scriptsize\textcolor{green!70!black}{(+5.1)}} & 69.8 {\scriptsize\textcolor{green!70!black}{(+8.1)}} \\
Polar \cite{PolarTransformer} & 94.7 {\scriptsize\textcolor{green!70!black}{(+4.5)}} & 67.6 {\scriptsize\textcolor{green!70!black}{(+5.9)}} \\
LogPolar \cite{PolarTransformer} & 93.8 {\scriptsize\textcolor{green!70!black}{(+3.6)}} & 65.6 {\scriptsize\textcolor{green!70!black}{(+3.9)}} \\
Diffeo-STN \cite{Detlefsen2018} & 94.9 {\scriptsize\textcolor{green!70!black}{(+4.7)}} & 66.8 {\scriptsize\textcolor{green!70!black}{(+5.1)}} \\
P-STN (Gamma) \cite{Schwobel2022} & 93.7 {\scriptsize\textcolor{green!70!black}{(+3.5)}} & 65.7 {\scriptsize\textcolor{green!70!black}{(+4.0)}} \\
P-STN (Gauss) & 91.3 {\scriptsize\textcolor{green!70!black}{(+1.1)}} & 67.3 {\scriptsize\textcolor{green!70!black}{(+5.6)}} \\
Ours (Gamma) & \textbf{96.5 {\scriptsize\textcolor{green!70!black}{(+6.3)}}} & 72.5 {\scriptsize\textcolor{green!70!black}{(+10.8)}} \\
Ours (Gauss) & 95.4 {\scriptsize\textcolor{green!70!black}{(+5.2)}} & \textbf{\text{72.8 {\scriptsize\textcolor{green!70!black}{(+11.1)}}}} \\
\bottomrule
\end{tabular}%
\vspace{-0.8cm}
\end{wraptable}

\paragraph{Setup}
Ecuador-Moth \cite{Ecuador} comprises 1445 samples and 675 classes and EU-Moth \cite{EU} has 1650 samples and 200 classes.
We follow existing protocols for both moth benchmarks \cite{Saccader,Boostrapping} and use a Swin-Base backbone \cite{Swin} with pre-trained weights from ImageNet-21k \cite{deng2009imagenet}.
All images of moths are top-down captures with rotation and scaling as relevant transformations.
Rotation to account for arbitrary orientation under capture, and scaling to handle natural size variation across individuals. 
We additionally include shearings as they approximate perspective distortions induced by camera angle. 
We evaluate robustness using an augmented test set \cite{Hendrycks2019Robust} using roto-scaling with $\theta \in [0, 2\pi)$ and $s \in [0.75, 1]$. 

\paragraph{Fine-Grained Classification and Robustness}
Here we examine whether models can be both robust on the augmented test set while maintaining high performance on the vanilla test set.
The results are shown in \cref{fig:moth_performance}.
We trained the Swin-Base classification backbone on the vanilla datasets and on a spatially augmented training set (random shearing, scaling, and rotation).  
Augmented training is one of the core baselines when modelling robust functions against $\mathrm{Aff}_2(\mathbb{R})$-perturbations as it usually brings high performance gains and is hard to outperform \cite{ITS}.
Along this baseline, we include the vanilla STN \cite{STN} regressing $\mathbf{M}$ directly and a version that regresses $\mathcal{T}$ with separate heads \cite{Chen2021} (Head-STN).
We include the Diffeomorphic-STN by \cite{Detlefsen2018}, where we implemented a diffeomorphic transformation using matrix exponential of the skew-symmetric $\mathbf{M}$. 
Another line of approaches was introduced by \cite{PolarTransformer}, where $\mathbf{x}$ is transformed to polar or log-polar space to extract features.
Convolution in log-polar space is equivariant to rotations and scaling, hence suitable to regress $\mathbf{M}$.
We include P-STNs \cite{Schwobel2022}, which we study using our Gauss prior and their Gamma hyperprior.

\begin{wraptable}{r}{0.58\textwidth}
    \vspace{-1.2cm}
    \centering
    \scriptsize
    \caption{
    Accuracy (Acc) and ROC AUC (AUC) on MedMNIST with statistics over 5 runs.
    Color encodes whether our approach consistently \colorbox{red!10}{\strut under-} or \colorbox{blue!10}{\strut overperforms} vanilla MedViT-S or results are \colorbox{black!10}{\strut mixed}.
    }
    \begin{tabular}{llll}
\toprule
Dataset & Model & Acc & AUC \\

\specialrule{0.4pt}{0pt}{0pt}
\rowcolor{blue!10}\raise\belowrulesep\hbox{\strut} & MedViT-S & $41.50$ & $0.697$ \\
\rowcolor{blue!10}RetinaMNIST & Ours (Gauss) & $52.90 \pm 0.63$ & $0.752 \pm 0.003$ \\
\rowcolor{blue!10}\lower\aboverulesep\hbox{\strut} & Ours (Gamma) & $\bm{55.35} \pm 0.58$ & $\bm{0.754} \pm 0.004$ \\

\specialrule{0.4pt}{0pt}{0pt}
\rowcolor{blue!10}\raise\belowrulesep\hbox{\strut} & MedViT-S & $78.80$ & $0.950$ \\
\rowcolor{blue!10}OCTMNIST & Ours (Gauss) & $75.30 \pm 0.00$ & $\bm{0.956} \pm 0.000$ \\
\rowcolor{blue!10}\lower\aboverulesep\hbox{\strut} & Ours (Gamma) & $\bm{83.20} \pm 0.39$ & $\bm{0.956} \pm 0.002$ \\

\specialrule{0.4pt}{0pt}{0pt}
\rowcolor{blue!10}\raise\belowrulesep\hbox{\strut} & MedViT-S & $68.91$ & $0.930$ \\
\rowcolor{blue!10}TissueMNIST & Ours (Gauss) & $69.83 \pm 0.00$ & $0.935 \pm 0.000$ \\
\rowcolor{blue!10}\lower\aboverulesep\hbox{\strut} & Ours (Gamma) & $\bm{72.47} \pm 0.04$ & $\bm{0.943} \pm 0.000$ \\

\specialrule{0.4pt}{0pt}{0pt}
\rowcolor{blue!10}\raise\belowrulesep\hbox{\strut} & MedViT-S & $91.36$ & $0.978$ \\
\rowcolor{blue!10}PathMNIST & Ours (Gauss) & $91.25 \pm 0.00$ & $0.977 \pm 0.000$ \\
\rowcolor{blue!10}\lower\aboverulesep\hbox{\strut} & Ours (Gamma) & $\bm{91.75} \pm 0.02$ & $\bm{0.986} \pm 0.000$ \\

\specialrule{0.4pt}{0pt}{0pt}
\rowcolor{blue!10}\raise\belowrulesep\hbox{\strut} & MedViT-S & $88.14$ & $0.963$ \\
\rowcolor{blue!10}PneumoniaMNIST & Ours (Gauss) & $88.04 \pm 0.09$ & $0.954 \pm 0.001$ \\
\rowcolor{blue!10}\lower\aboverulesep\hbox{\strut} & Ours (Gamma) & $\bm{93.21} \pm 0.18$ & $\bm{0.974} \pm 0.002$ \\

\specialrule{0.4pt}{0pt}{0pt}
\rowcolor{blue!10}\raise\belowrulesep\hbox{\strut} & MedViT-S & $76.61$ & $0.923$ \\
\rowcolor{blue!10}DermaMNIST & Ours (Gauss) & $74.26 \pm 0.04$ & $0.916 \pm 0.000$ \\
\rowcolor{blue!10}\lower\aboverulesep\hbox{\strut} & Ours (Gamma) & $\bm{77.67} \pm 0.43$ & $\bm{0.938} \pm 0.002$ \\

\specialrule{0.4pt}{0pt}{0pt}
\rowcolor{black!10}\raise\belowrulesep\hbox{\strut} & MedViT-S & $\bm{94.79}$ & $0.763$ \\
\rowcolor{black!10}ChestMNIST & Ours (Gauss) & $94.76 \pm 0.00$ & $0.754 \pm 0.000$ \\
\rowcolor{black!10}\lower\aboverulesep\hbox{\strut} & Ours (Gamma) & $94.75 \pm 0.00$ & $\bm{0.778} \pm 0.000$ \\
 
\specialrule{0.4pt}{0pt}{0pt}
\rowcolor{black!10}\raise\belowrulesep\hbox{\strut} & MedViT-S & $\bm{95.32}$ & $\bm{0.997}$ \\
\rowcolor{black!10}BloodMNIST & Ours (Gauss) & $94.84 \pm 0.05$ & $\bm{0.997} \pm 0.000$ \\
\rowcolor{black!10}\lower\aboverulesep\hbox{\strut} & Ours (Gamma) & $94.64 \pm 0.13$ & $\bm{0.997} \pm 0.000$ \\

\specialrule{0.4pt}{0pt}{0pt}
\rowcolor{black!10}\raise\belowrulesep\hbox{\strut} & MedViT-S & $93.10$ & $\bm{0.996}$ \\
\rowcolor{black!10}OrganCMNIST & Ours (Gauss) & $\bm{94.10} \pm 0.05$ & $\bm{0.996} \pm 0.000$ \\
\rowcolor{black!10}\lower\aboverulesep\hbox{\strut} & Ours (Gamma) & $91.77 \pm 0.07$ & $0.994 \pm 0.000$ \\

\specialrule{0.4pt}{0pt}{0pt}
\rowcolor{black!10}\raise\belowrulesep\hbox{\strut} & MedViT-S & $82.06$ & $\bm{0.976}$ \\
\rowcolor{black!10}OrganSMNIST & Ours (Gauss) & $\bm{82.81} \pm 0.01$ & $0.971 \pm 0.000$ \\
\rowcolor{black!10}\lower\aboverulesep\hbox{\strut} & Ours (Gamma) & $79.50 \pm 0.24$ & $0.972 \pm 0.000$ \\
 
\specialrule{0.4pt}{0pt}{0pt}
\rowcolor{black!10}\raise\belowrulesep\hbox{\strut} & MedViT-S & $\bm{95.49}$ & $\bm{0.998}$ \\
\rowcolor{black!10}OrganAMNIST & Ours (Gauss) & $94.99 \pm 0.00$ & $\bm{0.998} \pm 0.000$ \\
\rowcolor{black!10}\lower\aboverulesep\hbox{\strut} & Ours (Gamma) & $94.97 \pm 0.03$ & $0.997 \pm 0.000$ \\

\specialrule{0.4pt}{0pt}{0pt}
\rowcolor{red!10}\raise\belowrulesep\hbox{\strut} & MedViT-S & $\bm{92.31}$ & $\bm{0.916}$ \\
\rowcolor{red!10}BreastMNIST & Ours (Gauss) & $82.18 \pm 0.54$ & $0.890 \pm 0.001$ \\
\rowcolor{red!10}\lower\aboverulesep\hbox{\strut} & Ours (Gamma) & $84.74 \pm 0.54$ & $0.895 \pm 0.008$ \\

\specialrule{\heavyrulewidth}{0pt}{0pt}
\end{tabular}

    \label{tab:medmnist}
    \vspace{-1.2cm}
\end{wraptable}

All models are trained on augmented data, which is necessary as the rotoscale variance in both datasets is very minor.
Our model outperformed most baselines and is part of the Pareto front in both benchmarks.
Interestingly, all models fail to improve the top-1 performance on ECU.
On EU only the Head-STN and the augmented classifier are slightly outperforming our model on the vanilla test set.
We construct a third test set by augmenting $\mathcal{D}_{\text{test}}$ by random shearing first and then construct the orbit-space from before.
This limits the test set sizes to a reasonable level as the size increases with the order of the group when constructing orbit spaces.
The results are shown in \cref{tab:moths_shearing}.
We found very minor influence of the additional shearing to the performances of most tested baselines as well as our models.
The vanilla SwinB shows the most significant drop, which is expected.
We continue this study in the Supplementary Material.

\subsection{Medical Diagnostics}
\label{sec:sec:medmnist}

\paragraph{Setup}
We examine the predictive performance of MedViT-S \cite{MedViT} in its original configuration and in our framework (using rotation and scaling heads) on 12 2D datasets within MedMNISTv2 \cite{medMNIST}.
MedViT-S is the mid-sized model from the three models in the MedViT family, which are all hybrid neural networks, with both convolutional and attention layers.
To build our STN from it, we share the first MedViT-S layers up to the patch embedding of stage 2 with our localization encoder.
This encompasses the MedViT stem (4 convolutional layers), stage 1 (a patch embedding layer and 3 MedViT ECB blocks) and the patch embedding layer of stage 2.
We aligned our training procedure with that described in MedMNISTv2, similar to the original evaluation published with MedViT with minor changes.
For the initialization we use MedViT's ImageNet-1k pretrained weights (half frozen).
The pretrained weights are publicly available and referenced in the MedViT GitHub repository \cite{medvit_github}.
This fully includes the layers that are shared with the localization encoder (first two) in our framework.
We reduced the initial learning rate from 0.001 to 0.0001 to preserve the pretrained feature extraction and trained our models for only 50 instead of 100 epochs to reduce computational load.
To prevent overfitting, we used label smoothing of 0.1.
For a fair evaluation, in \Cref{tab:medmnist} we report new test-set results after training with these settings, both for MedViT-S in its original configuration and within our architectural framework.
We leverage the findings of the ablation in the preceding experiment.

\begin{wrapfigure}{r}{0.6\textwidth}
    \centering
    \vspace{-0.85cm}
    \includegraphics[width=1.\linewidth]{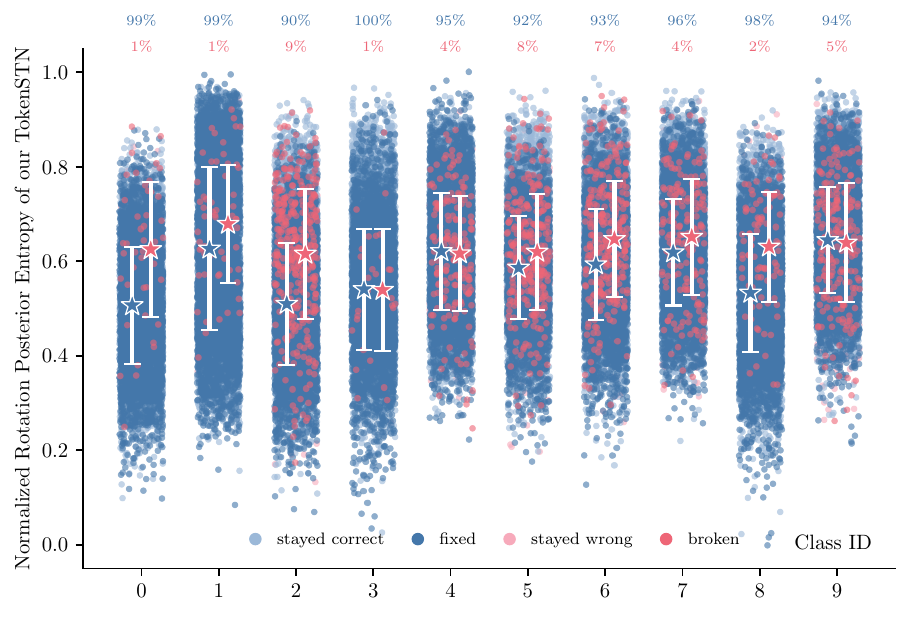}
    \caption{
        Normalized entropy of the predicted rotation posterior per RotMNIST class, using a frozen ImageNet-1k ViT-B/8 as the downstream classifier. 
        Each point is a test image, colored by how the rectification altered its prediction relative to the  uncorrected baseline.
        Stars mark the per-class mean entropy of the fixed and broken subsets, with error bars showing one standard deviation. 
        Percentages above each class give the conditional flip rates: fixed as a fraction of base-misclassified inputs, broken as a fraction of base-correct inputs. 
        Entropy is a monotone function of the predicted rotation standard deviation, so higher values indicate greater geometric uncertainty.
    }
    \label{fig:entropy_by_class_frozen}
    \vspace{-0.5cm}
\end{wrapfigure}

\vspace{-0.16cm}
\paragraph{Augmentation-free Classification}
We compare the predictive performance of MedViT-S in its original configuration (vanilla) as the downstream classifier for our STN.
The results in \Cref{tab:medmnist} show that our approach is almost always on par with vanilla MedViT-S or consistently outperforms it.
Although we noticed mode collapses (highly similar predictions for geometric primitives across samples), the positive results demonstrate stable fallback behavior.
This includes datasets where an improvement based on all predicted transformation parameters would not be expected, such as PneumoniaMNIST.
This dataset is made up of chest X-ray images of upright oriented patients with minimal rotational noise.
BreastMNIST is the only dataset in this experiment where both our models achieved lower accuracy and AUC than the baseline.
This can likely be attributed to the combination of a low number of samples (546 in training, the lowest in MedMNIST2D) and the present ultrasound image modality, which is sensitive to rotation, as one image side corresponds to the signal source.
At this number of samples, the downstream model can struggle to learn to be sufficiently robust against geometric perturbations, making it vulnerable to noise from the STN.

\subsection{Uncertainty of the Geometric Correction}
\label{sec:uncertainty}

We examine what the predicted posterior reveals about the reliability of our correction, and whether its uncertainty carries usable signal. 
We use a frozen ImageNet-1k~\cite{deng2009imagenet} pretrained ViT-B/8~\cite{ViT} and train only the STN, so that every prediction change is attributable to the geometric rectification alone rather than to classifier adaptation. 
The fine-tuned classifier reaches $98.87\%$ and $52.95\%$ top-1 test accuracy on the vanilla and the rotation-robustness test set, respectively.
With our TokenSTN, we boost the robustness to $96.10\%$ with only a marginal $2\%$ drop in vanilla performance.
\Cref{fig:entropy_by_class_frozen} reports the posterior entropy for each RotMNIST~\cite{RotMNIST} class. 
The percentages above each class give the conditional flip rates: among inputs the baseline misclassifies, the fraction recovered after rectification (fixed), and among inputs it classifies correctly, the fraction subsequently lost (broke). 
Across all classes the fixed rate dominates the broke rate by a large margin, recovering a median of 95.6\% (range 92--100\%) of previously misclassified inputs while corrupting at most 9\% of correct ones. 
For that worse case class (2) $4510/4991$ are corrected, while $372/4297$ are broken. 
The correction is thus strongly asymmetric in effect: it resolves many pose-induced errors at negligible risk to inputs that were already correct, and does so without any adaptation of the downstream model. 
Beyond its effect, the correction also exposes how uncertain it is. 
The posterior entropy is far from constant, spanning a wide range within every class rather than collapsing to its initialization, which confirms that the variational rotation head learns an input-dependent uncertainty over a named geometric quantity. 
We observe a consistent tendency for misclassified inputs to carry higher entropy.
We treat this uncertainty as an informative descriptor of the applied correction rather than a standalone failure detector, and leave a dedicated selective-prediction study to future work.

\subsection{Ablation Study}
\label{sec:ablation}

\paragraph{Component Ablations.}

\begin{wrapfigure}{r}{0.6\textwidth}
    \centering
    \vspace{-1cm}
    \includegraphics[width=1.\linewidth]{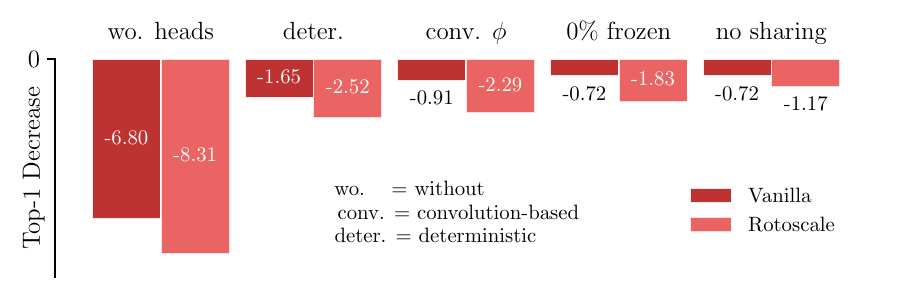}
    \caption{
    Top-1 test accuracy decrease on the vanilla and roto-scaled EU-Moth benchmark.
    These ablation results highlight the performance drops when disabling or replacing parts of our algorithm.
    }
    \label{fig:ablation}
    \vspace{-0.5cm}
\end{wrapfigure}

We study the performance impact of each key component of our STN through an ablation analysis.
We investigated the top-1 test accuracy gap between vanilla and roto-scaled versions of the EU-Moth \cite{EU} benchmark. 
Our results are presented in \cref{fig:ablation}.
Most significantly, we found that the decomposed regression approach in \cref{eq:bound_heads} (compared to direct regression of $\mathbf{M}$ as in \cite{STN}) yielded the largest performance gain.
The probabilistic regression of $\mathcal{T}$ in \cref{eq:sampling}, rather than deterministic regression, produced the second-largest improvement.
We also evaluated the impact of our token-based localization network and found that using a ConvNet operating on raw pixel space (as employed in nearly all related works) decreased performance.
We found sharing layers between $c$ and $\phi$ to improve performance as well as freezing early layers including the tokenizer. 
The performance drops observed when disabling or replacing components of our algorithm had greater impact on the roto-scaled test set than on the vanilla version. 
Our setup learns a more robust rectification, and since we evaluate on the roto-scaled test set, misclassification compounds exponentially across test images and affine transformation factors (see the Supplementary Material).

\begin{wrapfigure}{r}{0.55\textwidth}
    \centering
    \vspace{-0.3cm}
    \includegraphics[width=1.\linewidth]{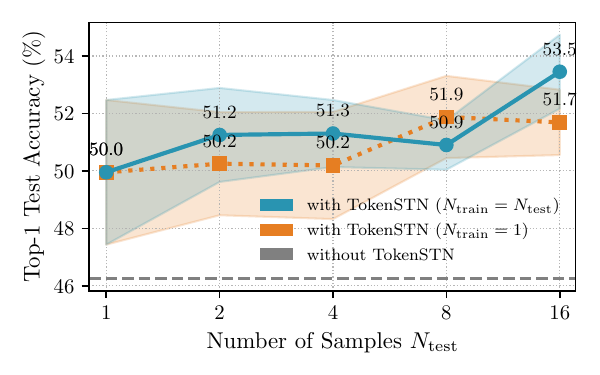}
    \caption{
        Average top-1 test accuracy over 5 seeded runs of the MedViT-S using our TokenSTN with different number of samples.
        This includes two modes: (blue) the number of samples is the same during training and testing, (orange) the number differs between training and test time.
        The gray curve is the baseline of the vanilla classifier.
    }
    \label{fig:num_samples}
    \vspace{-0.5cm}
\end{wrapfigure}

\paragraph{Parameter and Compute Efficiency.}
Our localization network $\phi$ reuses the shared encoder and adds only a shallow 2-layer transformer with lightweight heads, raising the trainable parameter count by roughly $2\%$ (under $1\mathrm{M}$). 
The only inference overhead is the Monte Carlo estimate of the predictive distribution, which scales linearly in $N$, as in comparable sampling-based methods~\cite{Schwobel2022, ITS, ITS2}. 
Since accuracy increases monotonically with $N$ (\cref{fig:num_samples}), a single trained model serves a range of compute budgets without retraining. 
More details in the Supplementary Material.

\section{Conclusion} \label{sec:conclusion}
We revisited STNs as a flexible, data-driven alternative to strictly invariant architectures for handling geometric variability in image classification. 
We extend STNs to token-based transformers and stabilize training by regressing constrained affine primitives with shared backbone representations.
Rectifications are measurable as a key part of the framework and thus require no post-hoc explanation.
Our framework maintains the ability to model general affine transformations without significant computational overhead.
Extensive experiments on fine-grained biodiversity benchmarks and diverse medical imaging tasks demonstrate that our approach consistently improves robustness to spatial transformations without sacrificing performance on canonical inputs. 
Overall, token-based STNs emerge as an effective mechanism for geometric robustness, combining structured geometric regression with the adaptability of modern vision models.

Mode collapses remain an open challenge for STNs, including ours. 
Objectives from JEPA training offer a principled path towards preserving variance, thereby potentially yielding more reliably meaningful corrections and clearer inspection.
Our framework also extends beyond the 2D planar transformations studied here.
Local transformations are one direction, via piecewise-linear affine maps or pixel-level velocity fields.
A step from 2D to 3D is another, for point cloud or mesh data. 
Finally, emphasizing downstream entropy minimization during training could sharpen automatic explanations. 
In integrated gradients \cite{pmlr-v70-sundararajan17a}, a more confident prediction on a high-entropy baseline could expand the attribution budget available to discriminative pixels.

\section*{Acknowledgements}
Johann Schmidt was funded by the ZIM (KK5540302MS4).
Tom Siegl was cofunded by the BMFTR (01IS22077, 01ED2507) and DFG CRC 1713/534829736.

\bibliographystyle{splncs04}
\bibliography{refs}

\newpage
\appendix
\section{Implementation Details}
\label{app:implementation}

\paragraph{Hardware.}
For each model configuration, NVIDIA H200 GPUs were utilized for about 24 (resp. 48) GPU hours to train for 50 (resp. 100) epochs and test on all MedMNIST datasets.
Our model based on MedViT-S used at most 71GiB GPU memory during the process, at a batch size of 128.

\paragraph{General Hyperparameters.}
We employ reflection padding to prevent boundary artifacts during resampling of the transformed image. 
We train all models over 100 epochs to ensure convergence, if not stated otherwise.
We freeze the first $50\%$ of parameters in the classifier (including shared layers).
In our regression heads, we use $\alpha_{\mathcal{T}} = 6$ and $\beta_{\mathcal{T}} = 4$.
Again following \cite{Schwobel2022} we use $N=1$ during training and $N=10$ during test-time.
In our regression bounding, we set $\lambda_\theta=1$ for full-orbit rotations, and $\lambda_s=\lambda_h=0.25$ and $\lambda_t=\tfrac{1}{8}$ to bound scaling, shearing and translation to reasonable ranges.
We use $\lambda=0.1$ in our loss to down-scale the KL loss term \cite{BetaVAE}.

\paragraph{Moth Experiments.}
For the SwinB experiments in our biodiversity monitoring experiments, we used the following settings:
We used the standard Swin configurations with $4 \times 4$ patch sizes, 8 attention heads, MLP ratio of 4, relative positional encoding, and GeLU \cite{GELU} activations. 
We use AdamW \cite{AdamW} for optimization with a cosine learning rate scheduler \cite{SGDR}.
Images are resized to $224 \times 224$ pixels and standardized using ImageNet statistics.
We apply label smoothing (0.1), dropout (20\% on the final encoder layer), weight decay (0.05), and gradient clipping at norm 1 for regularization.

\paragraph{MNIST Experiments.}
We used a ViT-B/8 \cite{ViT} pretrained on ImageNet1k \cite{ImageNet} zero-shot on RotMNIST (which is a rotation augmented MNIST test set).
We trained our TokenSTN with the frozen ViT-B/8 for $25$ epochs with uniform rotation augmentation. 
We used a learning rate of $0.0002$ and weight decay of $0.001$.
The TokenSTN uses only a rotation head (the rest is disabled) and a single sample for the inference.
No circular masking is used.

\newpage
\section{Further Experiments}
\label{app:experiments}

\subsection{Component Ablation Continued}
\label{app:ablation}

\begin{wrapfigure}{r}{0.6\textwidth}
    \centering
    \vspace{-0.8cm}
    \includegraphics[width=1.0\linewidth]{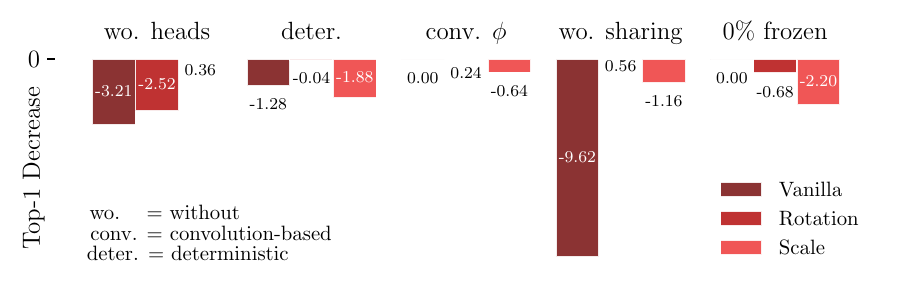}
    \caption{
    Component ablation using a MedViT-S on BreastMNIST.
    }
    \vspace{-0.8cm}
    \label{fig:ablation2}
\end{wrapfigure}

We confirm the findings in our ablation study (SwinB on EU-Moths) by conducting the same experiments for a MedViT-S on BreastMNIST (\cref{fig:ablation2}).
For a more detailed view, we tested on rotation and scaling test sets separately rather than the combined rotoscale performance. 
Again, none of the ablated components improved performance overall, confirming the relevance of all components in our pipeline.
While the findings on BreastMNIST largely agree with the findings on EU-Moths, we identified the following differences:
Without sharing any layers the vanilla performance had a significantly larger drop.
We conclude and highlight the importance of sharing encoder layers between the downstream model and the TokenSTN.
Without sharing, the model has to be trained from scratch without any prior feature detection abilities, which impacts performance.

\subsection{Number of Sample Ablation Continued}
\label{app:samples}

\begin{wrapfigure}{r}{0.5\textwidth}
    \centering
    \vspace{-0.8cm}
    \includegraphics[width=1.0\linewidth]{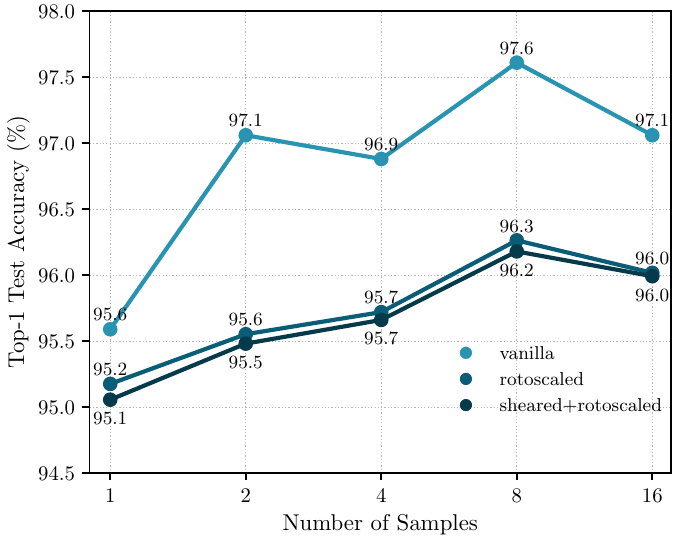}
    \caption{
    Top-1 test accuracy on the vanilla and augmented EU-Moth benchmark against different number of samples drawn from the posterior.
    Using the same number of samples during training and testing.
    }
    \vspace{-0.8cm}
    \label{fig:num_samples_moth}
\end{wrapfigure}

Building upon the findings on parameter and compute efficiency in our ablation study, we conducted an additional experiment on the EU-Moths dataset using the Swin-Base classifier to evaluate performance as a function of the sample size, $N$. 
As illustrated in \cref{fig:num_samples_moth}, all three test sets exhibit a clear positive trend. 
Performance improves as the number of samples increases, eventually plateauing and peaking at $N=8$ (in contrast to the peak at $N=16$ observed in our main ablation experiments). 
This asymptotic behavior is expected.
Consistent with the principle of diminishing returns, once a dataset is sufficiently represented, each additional sample yields progressively less novel information.

\subsection{Moth Experiments Continued}
\label{app:moths}

In \cref{tab:moth_performance} we provide the numerical values of our benchmark on the biodiversity monitoring task.
Alongside the seven STN baselines this includes the vanilla, training-time augmented, test-time augmented, trained and test-time canonicalization results.
Training-time augmentation trains the classifier by the expected transformation, while test-time augmentation (TTA, \cite{TIPooling}) uses the classifier trained on the vanilla training set and augments the test data instead.
By pooling over the predictions per transformed test sample induces invariance during test-time \cite{Puny2022}.
We use four random samples (drawn from the target group), so the result is a pseudo-invariant model.
Similar test-time sampling is performed by \cite{ITS}.
The difference is that samples are drawn as a regular grid over the group and the best overall sample is used for prediction. 
This results in a pseudo-canonicalization framework forming a tree search over multiple possible transformation levels.
We used two layers (rotation and scaling) and with eight samples each.
The canonicalization prior (CanPrior, \cite{Kaba2023, Mondal2023}) is limited to rotations here. 
We used an $SO(2)$-equivariant Group CNN \cite{GCNN} and a Steerable CNN \cite{Steerable} as backbones.
P-STN (Gauss) is the Schwöbel et al. \cite{Schwobel2022} model with the Gamma hyperprior swapped for a Gaussian.

\begin{table*}[t]
  \centering
  \caption{Average top-1 test accuracies on EU and ECU. Relative gains compared to vanilla baseline in parentheses. Extension of our benchmark on the biodiversity monitoring task.}
  \label{tab:moth_performance}
  \scriptsize
  \setlength{\tabcolsep}{6pt}
  \begin{tabular}{@{}lccccc @{}}
    \toprule
    \multirow{2}{*}{\textbf{Method}} & \multicolumn{2}{c}{\textbf{EU}} & \multicolumn{2}{c}{\textbf{ECU}} \\
    \cmidrule(lr){2-3} \cmidrule(lr){4-5}
    & Vanilla & RotoScale & Vanilla & RotoScale \\
    \midrule
    Vanilla & 96.3 & 90.6 & \textbf{76.2} & 62.8 \\
    Augmented & \textbf{97.1} {\scriptsize\textcolor{green!70!black}{$(+0.8)$}} & 95.1 {\scriptsize\textcolor{green!70!black}{$(+4.5)$}} & 73.0 {\scriptsize\textcolor{red}{$(-3.2)$}}   & 71.0 {\scriptsize\textcolor{green!70!black}{$(+8.2)$}} \\
    TTA \cite{TIPooling} & 96.3 {\scriptsize$(\pm0.0)$}                        & 92.6 {\scriptsize\textcolor{green!70!black}{$(+2.0)$}}  & \textbf{76.2} {\scriptsize$(\pm0.0)$}                       & 62.2 {\scriptsize\textcolor{red}{$(-0.6)$}} \\
    ITS \cite{ITS} & 94.4 {\scriptsize\textcolor{red}{$(-1.9)$}}   & 94.7 {\scriptsize\textcolor{green!70!black}{$(+4.1)$}} & 60.9 {\scriptsize\textcolor{red}{$(-15.3)$}} & 54.5 {\scriptsize\textcolor{red}{$(-7.3)$}} \\
    CanPrior (G) \cite{Mondal2023, GCNN} & 94.6 {\scriptsize\textcolor{red}{$(-1.7)$}} & 90.4 {\scriptsize\textcolor{red}{$(-0.2)$}} & 67.0 {\scriptsize\textcolor{red}{$(-9.2)$}} & 60.1 {\scriptsize\textcolor{red}{$(-2.7)$}}\\
    CanPrior (S) \cite{Mondal2023, Steerable} & 96.2 {\scriptsize\textcolor{red}{$(-0.1)$}} & 90.2 {\scriptsize\textcolor{red}{$(-0.4)$}} & 69.9 {\scriptsize\textcolor{red}{$(-6.3)$}} & 60.0 {\scriptsize\textcolor{red}{$(-2.8)$}} \\
    \midrule
    LogPolar \cite{PolarTransformer} & 94.9 {\scriptsize\textcolor{red}{$(-1.4)$}}   & 93.9 {\scriptsize\textcolor{green!70!black}{$(+3.3)$}}  & 67.8 {\scriptsize\textcolor{red}{$(-8.4)$}}   & 65.7 {\scriptsize\textcolor{green!70!black}{$(+2.9)$}} \\
    STN \cite{STN}      & 95.8 {\scriptsize\textcolor{red}{$(-0.5)$}}   & 94.9 {\scriptsize\textcolor{green!70!black}{$(+4.3)$}}  & 67.6 {\scriptsize\textcolor{red}{$(-8.6)$}}   & 65.5 {\scriptsize\textcolor{green!70!black}{$(+2.7)$}} \\
    P-STN (Gamma) \cite{Schwobel2022} & 95.5 {\scriptsize\textcolor{red}{$(-0.8)$}}   & 93.9 {\scriptsize\textcolor{green!70!black}{$(+3.3)$}}  & 69.0 {\scriptsize\textcolor{red}{$(-7.2)$}}   & 66.8 {\scriptsize\textcolor{green!70!black}{$(+4.0)$}} \\
    P-STN (Gauss) \cite{Schwobel2022} & 93.8 {\scriptsize\textcolor{red}{$(-2.5)$}}   & 91.2 {\scriptsize\textcolor{green!70!black}{$(+0.6)$}}  & 67.8 {\scriptsize\textcolor{red}{$(-8.4)$}}   & 67.2 {\scriptsize\textcolor{green!70!black}{$(+4.4)$}} \\
    Polar \cite{PolarTransformer} & 96.0 {\scriptsize\textcolor{red}{$(-0.3)$}}   & 94.7 {\scriptsize\textcolor{green!70!black}{$(+4.1)$}}  & 69.5 {\scriptsize\textcolor{red}{$(-6.7)$}}   & 67.8 {\scriptsize\textcolor{green!70!black}{$(+5.0)$}} \\
    Diffeo-STN \cite{Detlefsen2018} & 96.0 {\scriptsize\textcolor{red}{$(-0.3)$}}   & 94.8 {\scriptsize\textcolor{green!70!black}{$(+4.2)$}}  & 68.5 {\scriptsize\textcolor{red}{$(-7.7)$}}   & 64.0 {\scriptsize\textcolor{green!70!black}{$(+1.2)$}} \\
    Head-STN \cite{Chen2021} & \textbf{97.1} {\scriptsize\textcolor{green!70!black}{$(+0.8)$}}  & 95.4 {\scriptsize\textcolor{green!70!black}{$(+4.8)$}}  & 72.9 {\scriptsize\textcolor{red}{$(-3.3)$}}   & 70.1 {\scriptsize\textcolor{green!70!black}{$(+7.3)$}} \\
    \midrule
    \textbf{Ours (Gamma)} & 96.9 {\scriptsize\textcolor{green!70!black}{$(+0.6)$}}  & \textbf{96.7} {\scriptsize\textcolor{green!70!black}{$(+6.1)$}}  & 74.4 {\scriptsize\textcolor{red}{$(-1.8)$}}   & \textbf{73.0} {\scriptsize\textcolor{green!70!black}{$(+10.2)$}} \\
    \textbf{Ours (Gauss)} & 96.9 {\scriptsize\textcolor{green!70!black}{$(+0.6)$}}  & 95.4 {\scriptsize\textcolor{green!70!black}{$(+4.8)$}}  & 75.6 {\scriptsize\textcolor{red}{$(-0.6)$}}   & 72.7 {\scriptsize\textcolor{green!70!black}{$(+9.9)$}} \\
    \bottomrule
  \end{tabular}
\end{table*}

\subsection{Number of Parameters and FLOPS Continued}
\label{app:flops}

\begin{wrapfigure}{r}{0.6\textwidth}
    \vspace{-0.8cm}
    \centering
    \includegraphics[width=1.0\linewidth]{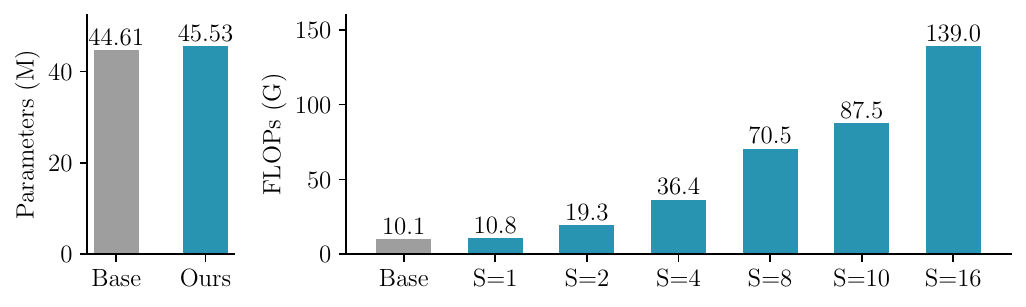}
    \caption{Number of parameters in millions and giga floating point operations per second (GFLOPS) using a MedViT-S classifier (base) and our STN with different number of samples.}
    \label{fig:flops}
    \vspace{-0.8cm}
\end{wrapfigure}

A central design goal of our framework is to add geometric robustness at minimal cost.
See \cref{fig:flops} for the parameter increase and FLOPS as a function of the number of samples $N$.
The light parameter increase results in a correspondingly small memory footprint, so the localization branch is effectively free relative to the backbone.
The remaining overhead stems entirely from the Monte Carlo estimate of the predictive distribution and scales linearly in the number of samples $N$, consistent with prior sampling based approaches \cite{Schwobel2022, ITS, ITS2}.
We regard this dependence as a strength rather than a drawback.
Since accuracy improves monotonically with $N$ (see the findings on parameter and compute efficiency in our ablation study), $N$ exposes a single, interpretable knob that trades compute for predictive performance.
This control is exercised purely at inference, so the same trained model can be deployed across a spectrum of compute regimes, from a fast single sample pass to a high accuracy multi sample evaluation, without any retraining or architectural change.
In settings with a fixed budget, $N$ can be set once to the largest affordable value, and in elastic or anytime settings it can be raised on demand.
Drawing more samples never degrades performance, so the tradeoff carries no downside beyond runtime.
In this sense, $N$ turns geometric robustness into a controllable resource rather than a fixed architectural cost.

\section{Decomposition}
\label{app:theory:decomposition}

\paragraph{A Note on Uniqueness}
The decomposition introduced above is unique as the degrees of freedom are equal.
$\mathbf{M}$ and its factored form has six degrees of freedom (parameter).
Reflection represents only a sign flag and is not counted here.
As this decomposition is unique, shearing and reflection along any line in 2D can be represented by the factored form.
Shearing and reflection are represented in canonical forms, i.e., we use shearing and reflection across the $x$-axis of the pixel grid.
Shearing and reflection across any line in 2D can be achieved by conjugation via rotation.\footnote{
For instance, using $\mathbf{R}_H^{-1} \mathbf{H} \mathbf{R}_H$, where $\mathbf{H}$ is the canonical shearing matrix and $\mathbf{R}_H$ a rotation matrix to the target line.
}
This conjugation does not appear explicitly in the factored form as its rotations are aggregated in the single rotation matrix $\mathbf{R}$.
This is possible as we only resample the image at the end, before the grid is transformed either by $\mathbf{M}$ or its factors such that previous grid points are mapped to new positions.
Due to the canonical shearing, the magnitude but not the orientation of the shear is factored. 

\section{Proofs}
\label{app:proofs}

\subsection{Proof of the Decomposition Lemma}
\label{proof:decomposition}

\begin{proof}
We seek a decomposition of the affine matrix $\mathbf{M}$ defined by the linear map $\mathbf{A} \in \mathrm{GL}_2(\mathbb{R})$ and translation $\mathbf{t} \in \mathbb{R}^2$.
Since matrix multiplication operates right-to-left, we first factor out the translation $\mathbf{t}$ to the left, leaving the linear part $\mathbf{A}$ in the upper-left block:
\begin{equation}
    \mathbf{M} = \underbrace{\begin{bNiceMatrix}[margin]
        1 & 0 & t_x \\
        0 & 1 & t_y \\
        0 & 0 & 1
    \end{bNiceMatrix}}_{\mathbf{T}}
    \begin{bNiceMatrix}[margin]
        \Block{2-2}{\mathbf{A}} & & 0 \\
        & & 0 \\
        0 & 0 & 1
    \end{bNiceMatrix}.
\end{equation}
As $\mathbf{A} \in \mathrm{GL}_2(\mathbb{R})$ is invertible, we decompose the linear block using \emph{QR decomposition} \cite[Chapter 5]{Golub1996}, which uniquely factors $\mathbf{A}$ into an orthogonal matrix $\mathbf{Q} \in \mathrm{O}(2)$ and an upper triangular matrix $\mathbf{K}$ with positive diagonal entries, such that $\mathbf{A} = \mathbf{Q}\mathbf{K}$.\footnote{Remark: The QL decomposition is a viable alternative resulting in the same outcome.}
The upper triangular matrix $\mathbf{K}$ captures the distortion. 
We decompose its affine form into a shear matrix $\mathbf{H}$ and a scaling matrix $\mathbf{S}$:
\begin{align}
    \underbrace{\begin{bNiceMatrix}[margin]
        k_{11} & k_{12} & 0 \\
        0 & k_{22} & 0 \\
        0 & 0 & 1
    \end{bNiceMatrix}}_{\mathbf{K}}
    &=
    \underbrace{\begin{bNiceMatrix}[margin]
        1 & h & 0 \\
        0 & 1 & 0 \\
        0 & 0 & 1
    \end{bNiceMatrix}}_{\mathbf{H}}
    \cdot
    \underbrace{\begin{bNiceMatrix}[margin]
        s_x & 0 & 0 \\
        0 & s_y & 0 \\
        0 & 0 & 1
    \end{bNiceMatrix}}_{\mathbf{S}} \nonumber \\
    &=
    \begin{bNiceMatrix}[margin]
        s_x & h s_y & 0 \\
        0 & s_y & 0 \\
        0 & 0 & 1
    \end{bNiceMatrix}.
\end{align}
Solving the system $k_{11}=s_x, k_{22}=s_y, k_{12}=h s_y$ yields the unique solution $s_x = k_{11}$, $s_y = k_{22}$, and $h = k_{12}/s_y$. 
Since $\mathbf{K}$ has only one off-diagonal degree of freedom, the single parameter $h$ fully characterizes the shear relative to the scaling axes.
The orthogonal matrix $\mathbf{Q}$ captures the isometry. 
Its determinant is $\det(\mathbf{Q}) = \det(\mathbf{A})/\det(\mathbf{K}) = \pm 1$. 
We isolate the rotation $\mathbf{R} \in \mathrm{SO}(2)$ and reflection $\mathbf{F}$ as follows:
\begin{itemize}
    \item[(i)] If $\det(\mathbf{A}) > 0$, then $\det(\mathbf{Q}) = 1$.
    We set the reflection parameter $r = 1$ (hence $\mathbf{F} = \mathbf{I}$) and $\mathbf{R} = \mathbf{Q}$.
    \item[(ii)] If $\det(\mathbf{A}) < 0$, then $\det(\mathbf{Q}) = -1$, meaning $\mathbf{Q}$ is a reflection-rotation. 
    By factoring out a canonical reflection $\mathbf{F}$ (where $r = -1$), we define $\mathbf{R} = \mathbf{F}^{-1}\mathbf{Q}$. 
    Since $\det(\mathbf{R}) = \det(\mathbf{F})\det(\mathbf{Q}) = (-1)(-1) = 1$, $\mathbf{R}$ is strictly a proper rotation:
    \begin{equation}
        \begin{bNiceMatrix}[margin]
            \Block{2-2}{\mathbf{Q}} & & 0 \\
            & & 0 \\
            0 & 0 & 1
        \end{bNiceMatrix}
        =
        \underbrace{\begin{bNiceMatrix}[margin]
            1 & 0 & 0 \\
            0 & -1 & 0 \\
            0 & 0 & 1
        \end{bNiceMatrix}}_{\mathbf{F}}
        \cdot
        \underbrace{\begin{bNiceMatrix}[margin]
            q_{11} & q_{12} & 0 \\
            -q_{21} & -q_{22} & 0 \\
            0 & 0 & 1
        \end{bNiceMatrix}}_{\mathbf{R}}.
    \end{equation}
\end{itemize}
Thus, one reflection parameter $r \in \{1, -1\}$ and one rotation angle $\theta$ suffice to represent any $\mathbf{Q} \in \mathrm{O}(2)$. 
Combining these factors yields $\mathbf{M} = \mathbf{T} \mathbf{F} \mathbf{R} \mathbf{H} \mathbf{S}$, which concludes the proof.
\end{proof}

\end{document}